\documentclass[conference]{IEEEtran}
\IEEEoverridecommandlockouts
\usepackage{cite}
\usepackage{amsmath,amssymb,amsfonts}
\usepackage{algorithmic}
\usepackage{graphicx}
\usepackage{textcomp}
\usepackage{xcolor}
\usepackage{orcidlink}
\usepackage{comment}
\usepackage{subcaption}
\usepackage{booktabs}
\usepackage{multirow}
\def\BibTeX{{\rm B\kern-.05em{\sc i\kern-.025em b}\kern-.08em
    T\kern-.1667em\lower.7ex\hbox{E}\kern-.125emX}}
\begin{document}

\title{Resource-Efficient 3D Shape and Orientation Estimation of Road Users using Roadside RGB Camera and 3D Radar}
\author{
    \IEEEauthorblockN{
        Shiva Agrawal\orcidlink{0000-0001-8633-341X}*, 
        Savankumar Bhanderi\orcidlink{0000-0001-7257-6736}, and 
        Gordon Elger\orcidlink{0000-0002-7643-7327}
    }
    \IEEEauthorblockA{
        Institute for Innovative Mobility (IIMo) \\
        Technische Hochschule Ingolstadt, Germany \\
        \{shiva.agrawal, savankumar.bhanderi, gordon.elger\}@thi.de \\
        *Corresponding Author: shiva.agrawal@thi.de, agrawalshivaa@gmail.com 
    }
}

\maketitle

\begin{abstract}
End-to-end deep learning pipelines for camera-radar 3D object detection are well established in driver assistance systems and autonomous vehicles; however, adapting them to roadside-mounted infrastructure presents significant challenges. On vehicles, sensor heights, pitch angles, and relative geometries remain largely uniform, allowing end-to-end models to generalize effectively. On the contrary, roadside sensor installations vary drastically in height, tilt, and field of view across locations, rendering end-to-end models prone to failure and difficult to adapt without massive location-specific datasets. Hence, instead of designing a complex end-to-end pipeline for direct 3D bounding box estimation, a low-cost, resource-efficient two-stage approach is proposed for real-time 3D shape and orientation estimation. The first stage utilizes any off-the-shelf, state-of-the-art method to extract 2D camera bounding boxes and associated 3D radar points for each road user. The proposed second stage then processes these inputs to construct 3D oriented bounding boxes where radar data determines 3D position, predefined templates define size for each road user, and a tiny neural network predicts a site-specific and optimized yaw angle. The algorithm is successfully validated at a curved road with a junction for pedestrians, bicycles, and cars, where it achieves $25$ frames per second on an NVIDIA Jetson ORIN for real-time applications.
\end{abstract}


\section{Introduction} \label{sec_introduction}

RGB (red, green, blue) cameras and 3D automotive radar sensors are widely used for precise road user detection and tracking in advanced driver assistance systems (ADAS) and autonomous vehicle development \cite{survey_paper_1, survey_paper_2}. Recently, with advancements in intelligent transportation systems (ITS), their deployment in roadside infrastructure setups has gained popularity for road user detection \cite{shiva_vehits_2022, in2lab_shiva, infra_low_cr_fusion_paper_1, infra_low_cr_fusion_paper_2}. These sensors operate based on different principles, providing complementary information that, when fused, enhances the overall accuracy and robustness of detection. However, due to the sparse nature of radar point clouds, fusion with camera data is often limited to detecting road users as point objects associated with categories and speed \cite{rc_high_level_vehicle_1}. Such point object detection identifies only sparse locations without shape or orientation details, limiting its practical usefulness. In contrast, representing road users as 3D oriented bounding boxes offers a comprehensive spatial understanding by capturing object size, position, and orientation, which are essential for accurate localization and tracking. 

Extensive research exists on direct 3D object detection using lidar sensors \cite{lidar_3d_object_detection_survey_1, lidar_3d_object_detection_survey_2}, as their high resolution enables precise 3D shape reconstruction. However, lidar’s high cost limits its practicality for intelligent transportation systems (ITS) that require widespread roadside deployment. Monocular camera-based 3D detection methods \cite{mono_camera_3d_object_detection_infra_1,mono_camera_3d_object_detection_infra_3,mono_camera_3d_object_detection_infra_4} are available but rely heavily on deep learning, demanding significant computational resources and large volumes of lidar-labelled data, which complicates real-time operation and development. Similarly, end-to-end deep learning-based fusion algorithms combining RGB cameras and radar \cite{cr_low_only_cars_3d_output_paper_1, cr_low_vehicle_3d_output_paper_1,  vehicle_3d_fusion_he2023concs, cr_low_vehicle_3d_output_paper_4} to predict 3D bounding boxes or bird’s-eye-view outputs depend heavily on 3D labels for training, further increasing complexity and cost.

On vehicles, sensor heights, pitch angles, and relative geometries remain largely uniform across different models, allowing end-to-end deep learning architectures to generalize effectively. In contrast, roadside camera and radar installations vary drastically in height, tilt, and field of view across locations. Consequently, an end-to-end fusion model trained on a specific roadside setup often fails at a new location because it learns feature mappings implicitly tied to that site's exact spatial configuration and 3D ground truth. This heavy reliance on site-specific geometry fundamentally restricts the generalizability of end-to-end models. Therefore, rather than forcing a single end-to-end fusion model to generalize across diverse environments, which is both costly and time-consuming, a more practical engineering solution is to design lightweight algorithms that quickly adapt to new sensor mountings. By shifting the focus toward rapid, location-specific adaptation, accurate and real-time 3D perception of road users can be reliably achieved across varying roadside infrastructure. 

Hence, a low-cost, resource-efficient two-stage approach
is proposed for real-time 3D shape and orientation estimation using a roadside-mounted RGB camera and 3D radar sensor. The first stage utilizes any off-the-shelf, state-of-the-art method
to extract 2D camera bounding boxes and associated 3D radar
points for each road user. The proposed second stage then
processes these inputs to construct 3D oriented bounding boxes
where radar data determines 3D position, predefined templates
define size for each road user, and a tiny neural network predicts
yaw angle. This decoupled architecture enables rapid, data-minimal deployment across diverse roadside sensor setups. The proposed methodology is validated for pedestrians, bicycles, and cars at a curved junction with a traffic light crossing.

\subsection{Key Contributions}
The main contributions of this work are summarized as follows:
\begin{enumerate}
    \item A resource-efficient 3D shape and orientation estimation methodology is developed, leveraging 2D camera bounding boxes and associated 3D radar points.
    \item Extensive ablation experiments are performed to establish optimal 3D size templates for pedestrians, cyclists, and cars.
    \item Quantitative validation against lidar ground truth is conducted to evaluate estimation accuracy in terms of 3D position, 3D size, and yaw angle.
    \item Comparative benchmarking against a camera-only homography 
\end{enumerate}
\section{Related Work} \label{sec_related_work}

Camera-based methods demonstrate that 3D bounding boxes can be inferred from monocular images using geometry and prior physical knowledge. The approach in \cite{C01} projects the lower contour of segmentation masks onto the ground plane via camera calibration to estimate a vehicle's footprint, from which height and orientation are derived geometrically. Similarly, \cite{C02, C03} reconstruct 3D boxes by combining camera calibration, vanishing points, and statistical vehicle dimensions, while \cite{C04} introduces a lightweight network for estimating object orientation and scale. The work in \cite{C05} further employs known-size templates of various road users alongside road geometry to transform 2D detections into 3D boxes. However, due to inherent limitations in direct depth estimation from monocular images, camera-only methods suffer from relatively low positional accuracy, which deteriorates significantly as distance from the sensor increases.

Radar sensors offer direct range and velocity measurements, making them ideal for precise spatial localization; however, processing radar data introduces unique algorithmic challenges due to severe point cloud sparsity, multipath reflections, and angular noise. Early short-range radar systems relied on beam-forming models \cite{R01}, though these were restricted to small sensing radii (e.g., within 5\,m). To improve spatial understanding at longer distances, high-resolution radar approaches \cite{R02} employ classical geometric features, such as applying the Hough transform on radar point reflections to extract vehicle orientation. Multi-radar networks and hybrid deep learning setups \cite{R03} further extend detection range and point density, yet struggle with false positive detections caused by clutter. Additionally, traditional geometric strategies like convex hull extraction and L-shape fitting adapted from lidar processing \cite{R04}, as well as template matching techniques \cite{R05}, enhance interpretability but remain tailored exclusively to large, highly reflective objects like cars.

While lidar-based architectures \cite{L01, L02, L03} achieve high 3D localization accuracy by processing dense spatial point clouds, these algorithms rely on fine spatial structures that do not translate directly to the significantly sparser and noisier distribution of radar targets.

To address these limitations without resorting to data-heavy end-to-end deep learning networks, this work strategically combines camera-based geometric projection and template-matching techniques with sparse radar point data to achieve accurate 3D positioning and sizing for road users. However, because classical geometric approaches struggle to reliably estimate yaw angles for bicycles and cars navigating curved roads and complex junctions, a tiny neural network optimized for site-specific performance is deployed in parallel. By coupling lightweight geometric priors with a compact model, the proposed framework achieves accurate, real-time 3D bounding box estimation with minimal computational overhead in a roadside camera-radar setup.
\section{Measurement setup} \label{sec:Measurement_setup}

\begin{figure}[ht]
\centering
\includegraphics[width=0.48\textwidth]{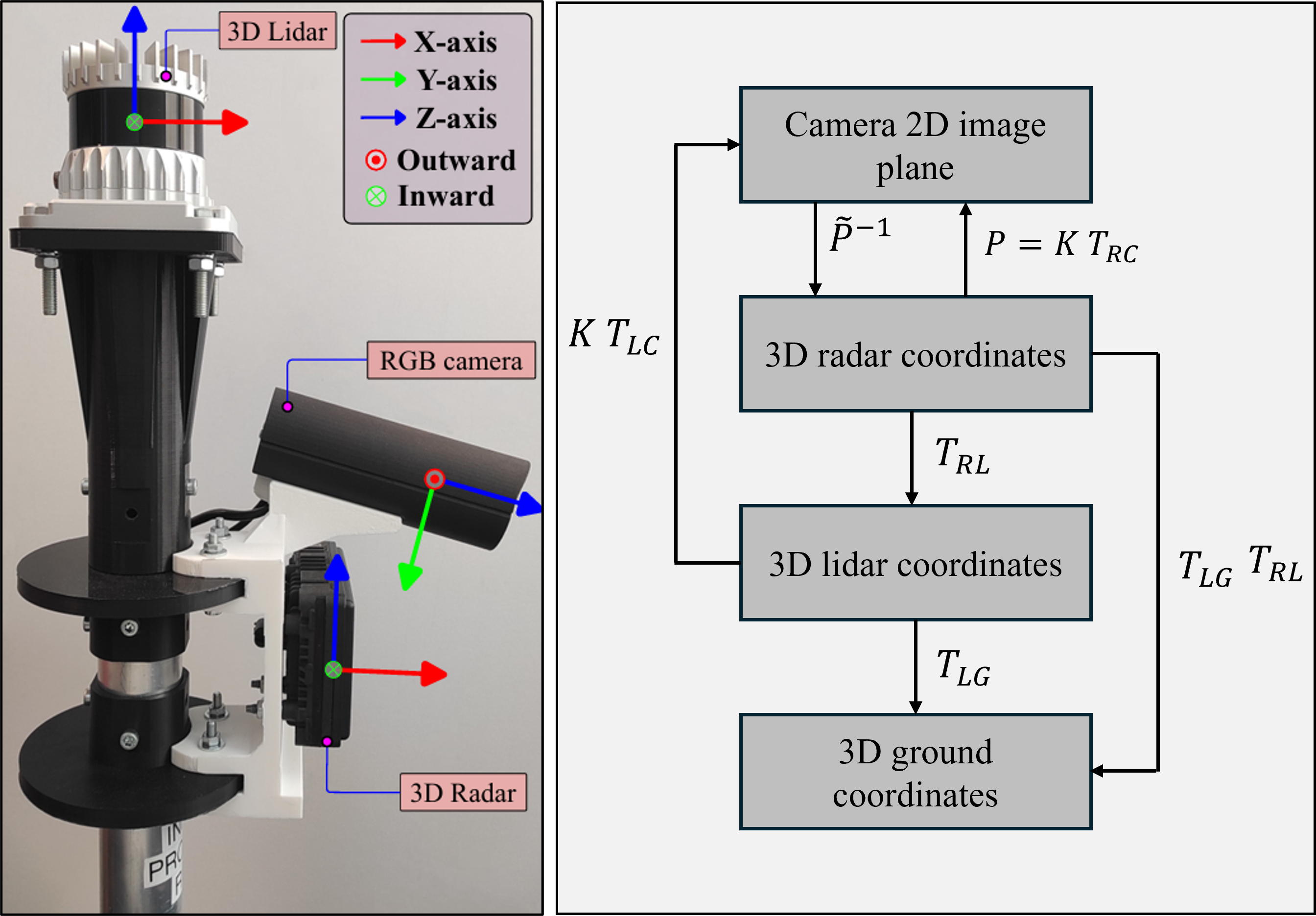}
\caption{Sensor setup and coordinate system (left) \cite{sensor_fusion_mdpi_shiva}, with corresponding coordinate transformations (right). $K$ denotes the camera intrinsic matrix; $T_{RC}, T_{LC}, T_{RL},$ and $T_{LG}$ represent the radar-to-camera, lidar-to-camera, radar-to-lidar, and lidar-to-ground extrinsic transformation matrices, respectively; $P$ and $\tilde{P}$ are the standard and homogeneous projection matrices.}
\label{fig:measurement_setup}
\end{figure}

The roadside multi-sensor setup used in this work is shown in Fig. \ref{fig:measurement_setup} (left). It is extended to approximately $3.5$ m from the ground using a laboratory tripod during data collection. It consists of an RGB mono camera (IDS UI-5260CP-C-HQ Rev.2) with an $8$ mm lens (Tamron M112FM08), a 3D automotive radar (ARS548), and a 3D automotive lidar sensor (OS1-64). 

Among them, the RGB mono camera and 3D automotive radar sensor are used for 3D shape and orientation estimation, and the 3D lidar sensor is used to validate the estimated output and for training the tiny neural network to predict the location-specific yaw angle for bicycles and cars. Details of the hardware, including sensors, electrical design, and mechanical design of this setup, are available in Sections $4$ and $5$ of \cite{shiva_ccis_springer_paper}. Robot operating system (ROS) $1$ approximate time synchronization is used to obtain the nearest time-synchronized frames from all three sensors for the recorded data at 10 Hz. Further, intrinsic calibration matrix ($K$) for the RGB camera is calculated using the state-of-the-art checker board method \cite{camera_intrinsic_checkerboard} and extrinsic calibration matrices such as radar-to-camera ($T_{RC}$), lidar-to-camera ($T_{LC}$), and radar-to-lidar ($T_{RL}$) are calculated using the methodology described in \cite{shiva_auto_calibration_paper}. The lidar-to-ground extrinsic matrix ($T_{LG}$) is derived by applying a vertical translation along the $Z$-axis, compensating for the physical mounting height of the sensor assembly. Utilizing these calibration matrices, the proposed framework performs various spatial coordinate transformations as illustrated in Fig.~\ref{fig:measurement_setup} (right).

\section{Methodology} \label{sec:methodology}

\begin{figure}[ht]
\centering
\includegraphics[width=0.48\textwidth]{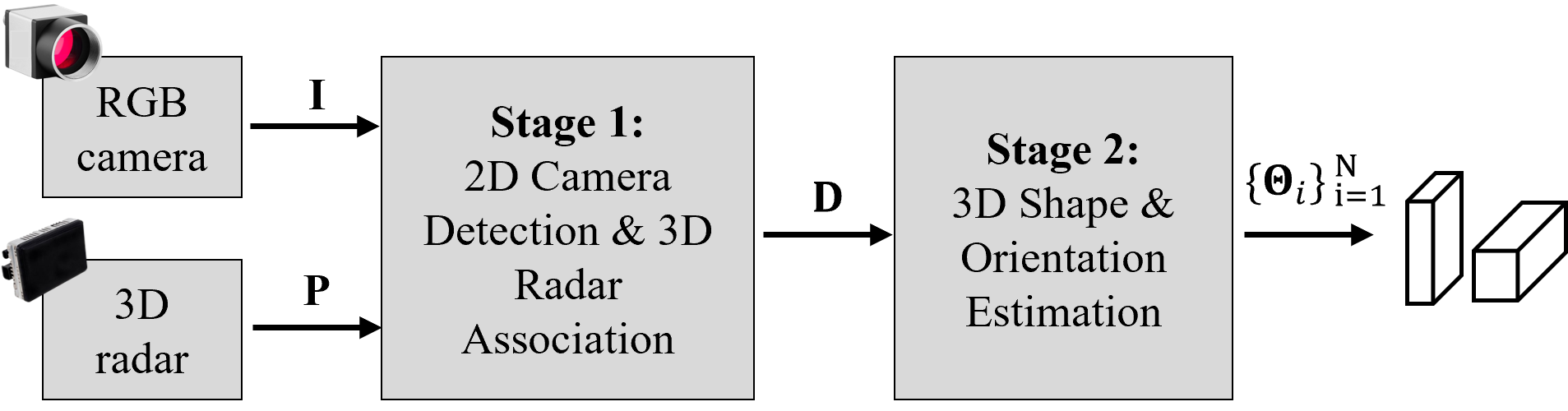}
\caption{Overview of the two-stage roadside 3D object detection framework. Inputs comprise a monocular image frame $\mathbf{I}$ and a sparse 3D radar point cloud frame $\mathbf{P}$. Stage 1 yields a set of $N$ road user instances $\mathbf{D} = \{(\mathbf{b}_i, c_i, s_i, \mathbf{P}_i)\}_{i=1}^N$, where $\mathbf{b}_i$ is the 2D bounding box, $c_i$ is the object category, $s_i$ is the detection confidence score, and $\mathbf{P}_i$ is the associated radar point subset. Stage 2 estimates the 3D Oriented Bounding Box (OBB) state vectors $\{\mathbf{\Theta}_i\}_{i=1}^N$ for all detected instances.}
\label{fig:two_stage_pipeline}
\end{figure}

The proposed two-stage framework for 3D road user detection using a roadside RGB camera and 3D radar is illustrated in Fig.~\ref{fig:two_stage_pipeline}. By decoupling instance association from 3D geometry estimation, this approach circumvents the need for complex, end-to-end deep learning models for direct 3D bounding box prediction. In Stage 1, processing is strictly restricted to 2D image detection and 3D radar point association, making it inherently independent of ground plane embeddings or location-specific camera-radar calibration priors. Subsequently, Stage 2 estimates the final 3D bounding boxes by deriving the 3D target positions directly from radar points, while leveraging predefined size templates and site-specific geometric constraints with minimal training data requirements. Consequently, rather than relying on complex end-to-end models that scale poorly to new sensor configurations in roadside sensor setups, Stage 2 focuses on finding an optimized, lightweight solution explicitly tailored for rapid, site-specific deployments.

\subsection{Problem Formulation}
Given a synchronized monocular image frame $\mathbf{I} \in \mathbb{R}^{H \times W \times 3}$ and a sparse radar point cloud $\mathbf{P} = \{\mathbf{p}_k \in \mathbb{R}^5\}_{k=1}^K$ containing 3D spatial coordinates, Doppler velocity, and Radar Cross Section (RCS), the first stage yields $N$ road user instances. For the $i$-th instance, the input comprises a 2D bounding box $\mathbf{b}_i \in \mathbb{R}^4$ in image space and an associated subset of 3D radar points $\mathbf{P}_i \subseteq \mathbf{P}$.

The objective is to estimate the 3D Oriented Bounding Box (OBB) parametrized by the state vector $\mathbf{\Theta}_i = \left( \mathbf{x}_i, \mathbf{d}_i, \psi_i, v_{r,i}, s_i, c_i \right)$, where $\mathbf{x}_i = (x_i, y_i, z_i)^\top \in \mathbb{R}^3$ represents the 3D bounding box centroid with respect to the roadside sensor setup in ground plane, $\mathbf{d}_i = (l_i, w_i, h_i)^\top \in \mathbb{R}^3$ denotes the 3D extent dimensions (length, width, height), $\psi_i \in \mathbb{SO}(2) \cong [-\pi, \pi]$ is the estimated yaw angle, $v_{r,i} \in \mathbb{R}$ is the radial Doppler velocity of the road user, $s_i \in [0, 1]$ is the detection confidence score (if obtained from stage-one), and $c_i \in \{1, \dots, C\}$ is the road user category index. In this work, pedestrians, bicycles, and cars are considered.

\subsection{Stage 1: 2D Camera Detection and 3D Radar Association}
In Stage 1, initial detections can be derived through state-of-the-art frame-wise object-level fusion \cite{rc_high_level_vehicle_1, cr_high_infra_3, cr_high_infra_4}, leveraging camera detectors such as YOLOv11 \cite{yolo11_ultralytics} alongside radar-based clustering or deep learning segmentation \cite{savan_radar_paper}. Alternatively, feature-level radar-camera fusion can be employed, as described in prior work \cite{sensor_fusion_mdpi_shiva}.  In our quantitative evaluation, Stage 1 uses YOLOv11  \cite{yolo11_ultralytics} (confidence threshold $s_i \ge 0.5$) for 2D bounding box extraction paired with DBSCAN spatial clustering for 3D radar point association. Unassociated or empty radar clusters default to monocular ground-plane homography depth estimates. The primary objective of this stage is to extract candidate 2D bounding boxes $\mathbf{b}_i$ from the monocular image $\mathbf{I}$ and isolate the corresponding 3D radar points $\mathbf{P}_i$ for each detected road user instance. The resulting output $\mathbf{D} = \{(\mathbf{b}_i, c_i, s_i, \mathbf{P}_i)\}_{i=1}^N$ serves as a lightweight, domain-agnostic representation containing all required multi-modal spatial cues for subsequent 3D estimation. 

\subsection{Stage 2: 3D Shape and Orientation Estimation}

Stage 2 receives the localized instance associations $\mathbf{D}$ from Stage 1 and constructs 3D bounding boxes, parameterized by the state vector $\mathbf{\Theta}_i = (x_i, y_i, z_i, l_i, w_i, h_i, \psi_i, v_{r,i}, s_i, c_i)$. In this proof-of-concept framework, three primary road user categories are evaluated: pedestrians, bicycles, and cars. 

The pipeline partitions 3D bounding box prediction into common preprocessing and category-specific processing. First, common preprocessing derives the 3D target centroid $\mathbf{x}_i = (x_i, y_i, z_i)^\top$ and mean Doppler velocity $v_{r,i}$ directly from the radar cluster $\mathbf{P}_i$, while extracting the height $h_i$ from the 2D image bounding box, alongside retaining confidence score $s_i$ and class $c_i$ from Stage 1 output. Subsequently, category-dependent processing estimates the remaining parameters: 2D ground extents $(l_i, w_i)$ are assigned using class-specific physical templates, and the yaw angle $\psi_i$ is predicted via lightweight neural networks for bicycles and cars. For pedestrians, $\psi_i$ is assigned a neutral $0^\circ$ orientation; since human physical profiles are spatially symmetric and lack a distinct front-to-rear directional axis, estimating a yaw angle is not required for frame-wise 3D object detection.

\subsubsection{Common Preprocessing}
The common preprocessing module computes the category-agnostic spatial and kinematic parameters, namely the 3D target centroid $\mathbf{x}_i$, radial Doppler velocity $v_{r,i}$, and height $h_i$. 

Let $\mathbf{P}_i = \{ (x^{(k)}, y^{(k)}, z^{(k)}, v_r^{(k)}) \}_{k=1}^{K_i}$ denote the set of $K_i$ radar points associated with target $i$ in ground coordinates (where RCS is omitted as it is unused in the geometric fitting process), and let $d_{\text{min}} = \min_k (x^{(k)})$ serve as the depth scaling factor $s$.

\textbf{3D Position Estimation:} The spatial centroid $\mathbf{x}_i$ = $(x_i, y_i, z_i)^\top$ is calculated as the mean over the radar point cluster $\mathbf{P}_i$ as per \eqref{eq:position_calc}.

\begin{equation} \label{eq:position_calc}
\mathbf{x}_i = \frac{1}{K_i} \sum_{k=1}^{K_i} \begin{bmatrix} x^{(k)} \\ y^{(k)} \\ z^{(k)} \end{bmatrix}
\end{equation}

\textbf{Radial Velocity Estimation:} Similarly, the target's mean radial Doppler velocity $v_{r,i}$ is computed across the cluster as per \eqref{eq:velocity_calc}.

\begin{equation} \label{eq:velocity_calc}
v_{r,i} = \frac{1}{K_i} \sum_{k=1}^{K_i} v_r^{(k)}
\end{equation}

\textbf{Height Estimation from 2D Bounding Box:} 
To estimate object height $h_i$, the four corners of the 2D image bounding box $\mathbf{b}_i = (u_{\min}, v_{\min}, u_{\max}, v_{\max})^\top$ are expressed in homogeneous pixel coordinates as $\mathbf{p}_m = [u_m, v_m, 1]^\top$ for $m \in \{1, 2, 3, 4\}$, corresponding to $(u_{\min}, v_{\min})$, $(u_{\max}, v_{\min})$, $(u_{\min}, v_{\max})$, and $(u_{\max}, v_{\max})$. 

Because mapping 2D image coordinates to 3D space is inherently unconstrained, recovering metric 3D position requires explicit depth information. In the presented pipeline, this metric depth scale $s$ is supplied directly by the associated radar cluster using its minimum longitudinal coordinate, such that $s = x_{\text{min}} = \min_k (x^{(k)})$. Given the camera projection matrix $\mathbf{P} = \mathbf{K} \cdot \mathbf{T}_{RC}$ (refer to Fig.~\ref{fig:measurement_setup} (right)), its augmented $4 \times 4$ homogeneous form $\tilde{\mathbf{P}}$ is constructed by appending the row vector $[0, 0, 0, 1]$. 

Each 2D pixel corner $\mathbf{p}_m$ is first inverse-projected into 3D radar space via $\tilde{\mathbf{P}}^{-1}$, and subsequently transformed into 3D ground coordinates $\mathbf{X}_{g,m} = [x_{g,m}, y_{g,m}, z_{g,m}, 1]^\top$ using the extrinsic transformations $\mathbf{T}_{RL}$ (radar-to-lidar) and $\mathbf{T}_{LG}$ (lidar-to-ground) as given in \eqref{eq:height_calc}.

\begin{equation} \label{eq:height_calc}
\mathbf{X}_{g,m} = \mathbf{T}_{LG} \cdot \mathbf{T}_{RL} \cdot \tilde{\mathbf{P}}^{-1} \cdot \left( s \cdot \begin{bmatrix} u_m \\ v_m \\ 1 \\ 1/s \end{bmatrix} \right)
\end{equation}

Finally, height $h_i$ is computed from the transformed 3D ground points by evaluating the absolute vertical difference along the $z$-axis between the top-edge corners ($m \in \{\text{top-left}, \text{top-right}\}$) and bottom-edge corners ($m \in \{\text{bottom-left}, \text{bottom-right}\}$):
\begin{equation}
h_i = | z_{g,\text{top}} - z_{g,\text{bot}} |
\end{equation}

\subsubsection{Category-dependent processing} 
While common preprocessing extracts primary spatial centroids, Doppler velocities, and heights, category-dependent refinement completes the state vector $\mathbf{\Theta}_i$ by predicting object orientation $\psi_i$ and 2D ground dimensions $(l_i, w_i)$ tailored to specific physical constraints of each road user class.

\textbf{Pedestrian:} 
For pedestrians, yaw angle estimation is omitted ($\psi_i = 0^\circ$) as human spatial symmetry makes orientation unnecessary for frame-wise detection. To determine 2D ground extents, a square footprint ($l_i = w_i$) is initially assigned. The initial side length is calculated analogously to the height estimation procedure (refer to \eqref{eq:height_calc}) by evaluating the absolute difference along the $y$-axis between the projected left and right horizontal corners of the 2D bounding box, as given in \eqref{eq:pedestrian_length_width_calc}:
\begin{equation} \label{eq:pedestrian_length_width_calc}
l_i = w_i = | y_{g,\text{right}} - y_{g,\text{left}} |
\end{equation}

However, because perspective projection causes dimensions derived directly from 2D image bounding boxes to vary with the camera viewing angle, these raw extents may not always be accurate. To compensate for this view-dependent distortion, distinct scaling factors $\alpha_{\text{ped}, l}$ (LSF: Length Scaling Factor) and $\alpha_{\text{ped}, w}$ (WSF: Width Scaling Factor) are applied to calibrate the final ground length and width as per \eqref{eq:ped_lw_multiplier}.

\begin{align} \label{eq:ped_lw_multiplier}
l_i &= \alpha_{\text{ped}, l} \cdot | y_{g,\text{right}} - y_{g,\text{left}} |, \\
w_i &= \alpha_{\text{ped}, w} \cdot | y_{g,\text{right}} - y_{g,\text{left}} |
\end{align}

where $\alpha_{\text{ped}, l}$ and $\alpha_{\text{ped}, w}$ are determined empirically through ablation experiments, as detailed in Section~\ref{sec:result_discussion}.

\textbf{Bicycle:}
Due to the sparse nature of the 3D radar sensor, most frames contain very few returns ($2$ to $4$ points) per bicycle, and frequently only a single point is detected. Consequently, experiments show that estimating a bicycle's 3D dimensions $(l_i, w_i)$ and yaw angle using spatial distribution methods such as convex hull, L-shape fitting, Doppler vector analysis, or 2D aspect-ratio heuristics is highly unreliable. Driven by point sparsity and image perspective distortion, these approaches cause severe frame-to-frame size variations and unstable orientation estimates.  

Hence, to ensure stable dimension estimation, the fitted length and width are bounded using pre-defined bicycle template limits $[l_{\text{bic},\min}, l_{\text{bic},\max}]$ and $[w_{\text{bic},\min}, w_{\text{bic},\max}]$, the values of which are optimized through ablation experiments (Section~\ref{sec:result_discussion}). For targets with $K_i \ge 2$ radar points, spatial extents $\Delta x_{\text{bic},i}$ and $\Delta y_{\text{bic},i}$ are measured across the cluster $\mathbf{P}_i$. The final bicycle length $l_{\text{bic},i}$ and width $w_{\text{bic},i}$ are then evaluated using \eqref{eq:bicycle_length} and \eqref{eq:bicycle_width} respectively. For $K_i = 1$, minimum values of length and width are used.

\begin{equation} \label{eq:bicycle_length}
l_{\text{bic},i} = \begin{cases} 
l_{\text{bic},\min}, & \text{if } \Delta x_{\text{bic},i} < l_{\text{bic},\min} \\
\Delta x_{\text{bic},i}, & \text{if } l_{\text{bic},\min} \le \Delta x_{\text{bic},i} \le l_{\text{bic},\max} \\
l_{\text{bic},\max}, & \text{if } \Delta x_{\text{bic},i} > l_{\text{bic},\max}
\end{cases}
\end{equation}

\begin{equation} \label{eq:bicycle_width}
w_{\text{bic},i} = \begin{cases} 
w_{\text{bic},\min}, & \text{if } \Delta y_{\text{bic},i} < w_{\text{bic},\min} \\
\Delta y_{\text{bic},i}, & \text{if } w_{\text{bic},\min} \le \Delta y_{\text{bic},i} \le w_{\text{bic},\max} \\
w_{\text{bic},\max}, & \text{if } \Delta y_{\text{bic},i} > w_{\text{bic},\max}
\end{cases}
\end{equation}

\begin{figure}[ht]
\centering
\includegraphics[width=0.40\textwidth]{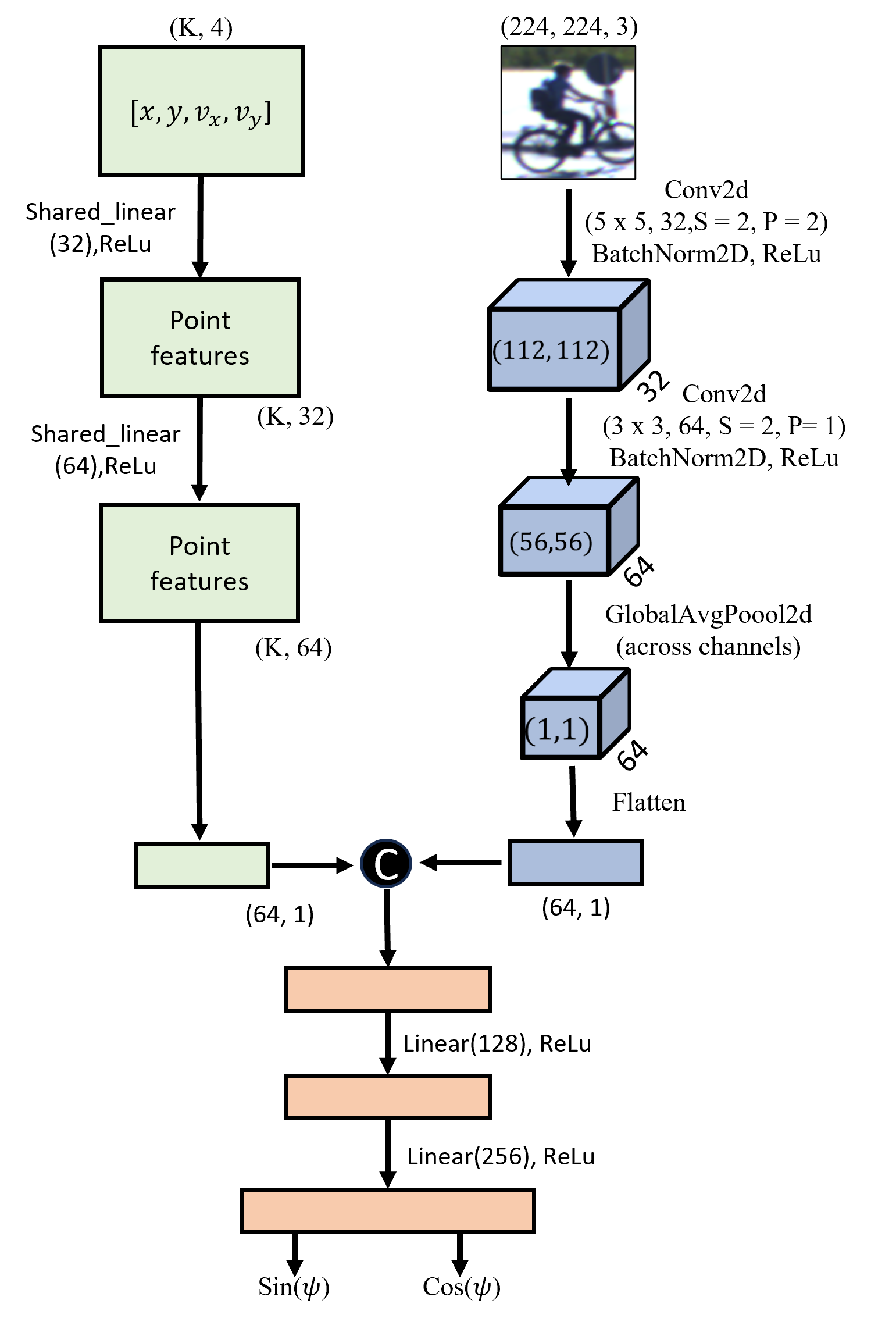}
\caption{A lightweight neural network architecture of \textit{YawPredictionNet} to estimate yaw angle ($\psi$) for bicycles.  It illustrates early feature extraction across radar and camera streams, feature concatenation, and continuous regression ($\sin\psi, \cos\psi$).}
\label{fig:yaw_prediction_net}
\end{figure}

For yaw angle estimation, a lightweight multi-modal neural network, \textit{YawPredictionNet} (as illustrated in Fig. \ref{fig:yaw_prediction_net}), is designed to jointly regress the sine and cosine components of the target orientation ($\sin\psi, \cos\psi$). The network processes two parallel input streams. Image ROI crops ($224 \times 224 \times 3$) are normalized via ImageNet statistics ($\boldsymbol{\mu} = [0.485, 0.456, 0.406]$, $\boldsymbol{\sigma} = [0.229, 0.224, 0.225]$) and  associated radar point fields $(x, y, v_x, v_y)$ are min-max normalized to $[-1, 1]$ using sensor bounds: $x \in [0, 80]\,\text{m}$, $y \in [-40, 40]\,\text{m}$, and $v_x, v_y \in [-25, 25]\,\text{m/s}$. Values of $v_x$ and $v_y$ for each radar point are calculated using ~\eqref{eq:radar_vx_vy}, where $v_r$ is the Doppler velocity, $\phi$ is the azimuth, and $\theta$ is the elevation angle measured directly by the radar sensor. 

\begin{equation} \label{eq:radar_vx_vy}
    \begin{aligned} 
    v_x = v_r \cos(\phi) \cos(\theta) \\
    v_y = v_r \cos(\phi) \sin(\theta)
\end{aligned}
\end{equation}

\textbf{Car:}
The 3D oriented bounding box of a car shares similar shape characteristics with a bicycle, differing primarily in its larger spatial dimensions. Consequently, the methods evaluated for estimating the length, width, and yaw angle of bicycles are directly extended to cars, yielding optimal performance. 

Specifically, the car length $l_{\text{car},i}$ and width $w_{\text{car},i}$ are evaluated using the same bounded piecewise formulations as in \eqref{eq:bicycle_length} and \eqref{eq:bicycle_width}, constrained within car-specific template limits $[l_{\text{car},\min}, l_{\text{car},\max}]$ and $[w_{\text{car},\min}, w_{\text{car},\max}]$. However, once $l_{\text{car},i}$ and $w_{\text{car},i}$ are determined, an additional centroid adjustment step is performed. The 3D bounding box centroid is shifted along the sensor line-of-sight towards the visible facets of the car to ensure precise spatial alignment with the sparse surface reflections.

For car yaw angle estimation, the same \textit{YawPredictionNet} architecture (Fig.~\ref{fig:yaw_prediction_net}) and input normalization protocols are adopted. To adjust the capacity for the car category, a modified variant is utilized wherein the final hidden linear layer ($256$ units) is omitted prior to regressing $(\sin\psi, \cos\psi)$.

\section{Experimental results and discussions} \label{sec:result_discussion}

Manually labeled 3D bounding boxes from the lidar sensor serve as the ground truth (GT) data in this work. Distinct datasets are dedicated to training and validating the \textit{YawPredictionNet} models. At the same time, a separate, independent set of $500$ frames containing $451$ persons, $360$ bicycles, and $299$ cars is reserved exclusively for validating the final 3D OBB outcome of the proposed methodology and is omitted from parameter tuning. To prevent temporal data leakage, dataset splits are created using sequence-disjoint segments rather than random frame sampling. This frame count was strategically chosen as manual 3D annotation of dense point clouds is exceptionally labor-intensive, yet the total count of over $1100$ well-distributed dynamic object instances across these frames provides a statistically robust and diverse evaluation set. At the selected location, the distribution of the road users is found to be in the range of $0$ to $50$ m from the sensor setup. 

\subsection{Training of YawPredictionNet}
Training and validation for \textit{YawPredictionNet} models for bicycles and cars is performed using AdamW (initial learning rate $10^{-3}$, batch size $1$, gradient clip $0.5$) with a \texttt{ReduceLROnPlateau} scheduler (patience 50 epochs) optimizing $L_{\text{total}} = L_{\text{MSE}}(\sin\psi) + L_{\text{MSE}}(\cos\psi)$. Class-specific parameters are configured as follows:
\begin{itemize}
    \item \textbf{Bicycle:} Trained for $400$ epochs ($262$ train / $66$ val samples) with weight decay $0.3$, requiring only $\mathbf{73,442}$ trainable parameters.
    \item \textbf{Car:} Trained for $100$ epochs ($329$ train / $76$ val samples) with weight decay $0.2$, requiring only $\mathbf{40,162}$ trainable parameters.
\end{itemize}

Owing to the lightweight architecture and minimal dataset size required for training, \textit{YawPredictionNet} is exceptionally deployment-friendly and easily adaptable to real-time embedded hardware. Furthermore, while lidar is utilized in this work, ground-truth yaw orientation can be conveniently sourced from alternative modalities.

\subsection{Ablation Experiments} 
Ablation experiments are conducted to determine optimal 3D bounding box dimensions (length, width, height) for pedestrians, bicycles, and cars. In each ablation experiment, the parameter change is evaluated across the dataset used for training the YawPredictionNet against the ground truth to compute the mean and standard deviation of errors (Table~\ref{table:person_ablation_eval} and ~\ref{table:bicycle_ablation_eval}) and absolute values (Table~\ref{table:car_ablation_eval}). 

Table~\ref{table:person_ablation_eval} shows the pedestrian parameter search across Length (LSF), Width (WSF), and Height (HSF) scaling factors. Starting from a square footprint assumption, LSF variations (Exp.~1--5) show that $\text{LSF}=1.75$ reduces length error to $(0.11, 0.09)$~m compared to $(0.44, 0.10)$~m at $\text{LSF}=1.00$. WSF variations (Exp.~6--7) confirm $\text{WSF}=1.25$ minimizes width error to $(0.07, 0.06)$~m. HSF variations (Exp.~8--9) verify that $\text{HSF}=1.00$ maintains the best height error at $(0.20, 0.11)$~m. Overall, Exp.~4 ($\text{LSF}=1.75$, $\text{WSF}=1.25$, $\text{HSF}=1.00$) yields the optimal configuration.

\def\arraystretch{1.2} 
\tabcolsep=3.0pt
\begin{table}[!h]
\centering
\caption{Ablation study of parameter variation on dimension errors for Pedestrians. LSF: Length Scaling Factor, WSF: Width Scaling Factor, HSF: Height Scaling Factor}
\label{table:person_ablation_eval}
\begin{tabular}{c c c c c c c}
\toprule \midrule
\multirow{2}{*}{\textbf{Exp.}} & \multicolumn{3}{c}{\textbf{Parameters varied}} & \multicolumn{3}{c}{\textbf{Errors (mean, std)}} \\ 
\cmidrule(lr){2-4} \cmidrule(lr){5-7}
& \textbf{LSF} & \textbf{WSF} & \textbf{HSF} & \textbf{Length} & \textbf{Width} & \textbf{Height} \\ 
\midrule \hline

1 & $1.25$ & $1.25$ & $1$    & $(0.27, 0.10)$ & $(0.07, 0.06)$ & $(0.20, 0.11)$ \\
2 & $1.00$ & $1.25$ & $1$    & $(0.44, 0.10)$ & $(0.07, 0.06)$ & $(0.20, 0.11)$ \\
3 & $1.50$ & $1.25$ & $1$    & $(0.13, 0.09)$ & $(0.07, 0.06)$ & $(0.20, 0.11)$ \\
4 & $\mathbf{1.75}$ & $\mathbf{1.25}$ & $\mathbf{1}$    & $\mathbf{(0.11, 0.09)}$ & $\mathbf{(0.07, 0.06)}$ & $\mathbf{(0.20, 0.11)}$ \\
5 & $2.00$ & $1.25$ & $1$    & $(0.23, 0.14)$ & $(0.07, 0.06)$ & $(0.20, 0.11)$ \\ \hline
6 & $1.75$ & $1.00$ & $1$    & $(0.11, 0.09)$ & $(0.20, 0.07)$ & $(0.20, 0.11)$ \\
7 & $1.75$ & $1.50$ & $1$    & $(0.11, 0.09)$ & $(0.15, 0.09)$ & $(0.20, 0.11)$ \\ \hline
8 & $1.75$ & $1.25$ & $1.25$ & $(0.11, 0.09)$ & $(0.07, 0.06)$ & $(0.26, 0.17)$ \\
9 & $1.75$ & $1.25$ & $0.75$ & $(0.11, 0.09)$ & $(0.07, 0.06)$ & $(0.58, 0.11)$ \\

\midrule
\bottomrule
\end{tabular}
\end{table}

Table~\ref{table:bicycle_ablation_eval} presents the parameter ablation for bicycles across length ranges, width ranges, and height scale factors (HSF). Evaluating length ranges (Exp.~1--5) shows that $2.0/2.4$~m achieves the minimum length error of $(0.10, 0.09)$~m. Width range variations (Exp.~6--8) confirm $0.8/1.0$~m minimizes width error to $(0.05, 0.04)$~m. HSF variations (Exp.~9--10) verify that scaling height increases error, confirming $\text{HSF}=1.00$ with a height error of $(0.10, 0.08)$~m. Thus, Exp.~6 yields the optimal parameter set.

\def\arraystretch{1.4} 
\tabcolsep=4pt
\begin{table}[!h]
\centering
\caption{Ablation study of parameter variation on dimension errors for bicycles. HSF: Height Scaling Factor}
\label{table:bicycle_ablation_eval}
\resizebox{\columnwidth}{!}{%
\begin{tabular}{c c c c c c c}
\toprule \midrule
\multirow{2}{*}{\textbf{Exp.}} & \multicolumn{3}{c}{\textbf{Parameters Varied}} & \multicolumn{3}{c}{\textbf{Errors (mean, std)}} \\ 
\cmidrule(lr){2-4} \cmidrule(lr){5-7}
& \textbf{Length (m)} & \textbf{Width (m)} & \textbf{HSF} & \textbf{Length} & \textbf{Width} & \textbf{Height} \\ 
& \textbf{min / max} & \textbf{min / max} & & & & \\
\midrule \hline

1  & $1.8 / 2.2$ & $0.7 / 1.0$ & $1.00$ & $(0.28, 0.12)$ & $(0.07, 0.05)$ & $(0.10, 0.08)$ \\
2  & $1.6 / 2.2$ & $0.7 / 1.0$ & $1.00$ & $(0.47, 0.12)$ & $(0.07, 0.05)$ & $(0.10, 0.08)$ \\
3  & $2.0 / 2.2$ & $0.7 / 1.0$ & $1.00$ & $(0.10, 0.09)$ & $(0.07, 0.05)$ & $(0.10, 0.08)$ \\
4  & $2.0 / 2.4$ & $0.7 / 1.0$ & $1.00$ & $(0.10, 0.09)$ & $(0.07, 0.05)$ & $(0.10, 0.08)$ \\
5  & $2.2 / 2.4$ & $0.7 / 1.0$ & $1.00$ & $(0.15, 0.07)$ & $(0.07, 0.05)$ & $(0.10, 0.08)$ \\ \hline
6  & $\mathbf{2.0 / 2.4}$ & $\mathbf{0.8 / 1.0}$ & $\mathbf{1.00}$ & $\mathbf{(0.10, 0.09)}$ & $\mathbf{(0.05, 0.04)}$ & $\mathbf{(0.10, 0.08)}$ \\
7  & $2.0 / 2.4$ & $0.9 / 1.0$ & $1.00$ & $(0.10, 0.09)$ & $(0.13, 0.05)$ & $(0.10, 0.08)$ \\
8  & $2.0 / 2.4$ & $0.9 / 1.1$ & $1.00$ & $(0.10, 0.09)$ & $(0.13, 0.06)$ & $(0.10, 0.08)$ \\ \hline
9  & $2.0 / 2.4$ & $0.9 / 1.1$ & $1.25$ & $(0.10, 0.09)$ & $(0.13, 0.06)$ & $(0.41, 0.16)$ \\
10 & $2.0 / 2.4$ & $0.9 / 1.1$ & $0.75$ & $(0.10, 0.09)$ & $(0.13, 0.06)$ & $(0.45, 0.10)$ \\

\midrule
\bottomrule
\end{tabular}%
}
\end{table}

Table~\ref{table:car_ablation_eval} presents the parameter ablation for cars. In our pipeline, initial $x_c$ and $y_c$ centroids are computed from the radar point cluster's spatial distribution; after fitting optimal length and width, the centroid is shifted to align the bounding box toward sensor-facing edges, linking dimension variations directly to positioning errors. Varying length limits (Exp.~1--8) shows $4.4/4.9$~m minimizes length error while keeping $x_c$ error low at $(0.38, 0.29)$~m. Width limit adjustments (Exp.~9--12) demonstrate that $1.8/2.0$~m achieves the optimal balance between width error $(0.15, 0.07)$~m and $y_c$ error $(0.55, 0.42)$~m—the latter absorbing the edge-alignment shift. Finally, HSF variations (Exp.~13--14) confirm that scaling height degrades accuracy, validating $\text{HSF}=1.00$. Overall, Exp.~12 yields the best trade-off across both dimension and centroid estimation.

While a manual search is used in this work to optimize the values of LSF, WSF, and HSF, alternative optimization techniques, such as Bayesian optimization, grid/random search, or multi-objective evolutionary algorithms (e.g., NSGA-II), could be applied, but they are out of scope of this work.

\def\arraystretch{1.2} 
\tabcolsep=4.2pt
\begin{table*}[!t]
\centering
\small
\caption{Ablation study of parameter variation on car dimension and centroid errors. HSF: Height Scaling Factor}
\label{table:car_ablation_eval}
\begin{tabular}{c c c c c c c c c c}
\toprule \midrule
\multirow{2}{*}{\textbf{Exp.}} & \multicolumn{3}{c}{\textbf{Parameters Varied}} & \multicolumn{6}{c}{\textbf{Errors (mean, std)}} \\ 
\cmidrule(lr){2-4} \cmidrule(lr){5-10}
& \textbf{Length (m)} & \textbf{Width (m)} & \textbf{HSF} & \textbf{$x_c$ (m)} & \textbf{Length (m)} & \textbf{$y_c$ (m)} & \textbf{Width (m)} & \textbf{$z_c$ (m)} & \textbf{Height (m)} \\ 
& \textbf{min / max} & \textbf{min / max} & & & & & & & \\
\midrule \hline

1  & $3.8 / 5.5$ & $1.7 / 2.2$ & $1.00$ & $(0.40, 0.39)$ & $(0.64, 0.22)$ & $(0.54, 0.42)$ & $(0.24, 0.08)$ & $(0.22, 0.06)$ & $(0.08, 0.09)$ \\
2  & $4.0 / 5.5$ & $1.7 / 2.2$ & $1.00$ & $(0.40, 0.38)$ & $(0.48, 0.22)$ & $(0.54, 0.42)$ & $(0.24, 0.08)$ & $(0.22, 0.06)$ & $(0.08, 0.09)$ \\
3  & $4.2 / 5.5$ & $1.7 / 2.2$ & $1.00$ & $(0.39, 0.33)$ & $(0.33, 0.21)$ & $(0.54, 0.42)$ & $(0.24, 0.08)$ & $(0.22, 0.06)$ & $(0.08, 0.09)$ \\
4  & $4.4 / 5.5$ & $1.7 / 2.2$ & $1.00$ & $(0.38, 0.29)$ & $(0.21, 0.06)$ & $(0.54, 0.42)$ & $(0.24, 0.08)$ & $(0.22, 0.06)$ & $(0.08, 0.09)$ \\
5  & $4.6 / 5.5$ & $1.7 / 2.2$ & $1.00$ & $(0.41, 0.30)$ & $(0.17, 0.24)$ & $(0.54, 0.42)$ & $(0.24, 0.08)$ & $(0.22, 0.06)$ & $(0.08, 0.09)$ \\ 
6  & $4.4 / 5.3$ & $1.7 / 2.2$ & $1.00$ & $(0.39, 0.29)$ & $(0.23, 0.11)$ & $(0.54, 0.42)$ & $(0.24, 0.08)$ & $(0.22, 0.06)$ & $(0.08, 0.09)$ \\
7  & $4.4 / 5.1$ & $1.7 / 2.2$ & $1.00$ & $(0.38, 0.29)$ & $(0.23, 0.08)$ & $(0.54, 0.42)$ & $(0.24, 0.08)$ & $(0.22, 0.06)$ & $(0.08, 0.09)$ \\
8  & $4.4 / 4.9$ & $1.7 / 2.2$ & $1.00$ & $(0.38, 0.29)$ & $(0.21, 0.05)$ & $(0.54, 0.42)$ & $(0.24, 0.08)$ & $(0.22, 0.06)$ & $(0.08, 0.09)$ \\ \hline
9  & $4.4 / 4.9$ & $1.5 / 2.2$ & $1.00$ & $(0.38, 0.29)$ & $(0.21, 0.05)$ & $(0.52, 0.40)$ & $(0.42, 0.10)$ & $(0.22, 0.06)$ & $(0.08, 0.09)$ \\
10 & $4.4 / 4.9$ & $1.9 / 2.2$ & $1.00$ & $(0.38, 0.29)$ & $(0.21, 0.05)$ & $(0.58, 0.44)$ & $(0.08, 0.04)$ & $(0.22, 0.06)$ & $(0.08, 0.09)$ \\
11 & $4.4 / 4.9$ & $1.8 / 2.2$ & $1.00$ & $(0.38, 0.29)$ & $(0.21, 0.05)$ & $(0.56, 0.43)$ & $(0.15, 0.07)$ & $(0.22, 0.06)$ & $(0.08, 0.09)$ \\
12 & $\mathbf{4.4 / 4.9}$ & $\mathbf{1.8 / 2.0}$ & $\mathbf{1.00}$ & $\mathbf{(0.38, 0.29)}$ & $\mathbf{(0.21, 0.05)}$ & $\mathbf{(0.55, 0.42)}$ & $\mathbf{(0.15, 0.07)}$ & $\mathbf{(0.22, 0.06)}$ & $\mathbf{(0.08, 0.09)}$ \\ \hline
13 & $4.4 / 4.9$ & $1.8 / 2.0$ & $1.25$ & $(0.38, 0.29)$ & $(0.21, 0.05)$ & $(0.55, 0.42)$ & $(0.15, 0.07)$ & $(0.07, 0.09)$ & $(0.39, 0.13)$ \\
14 & $4.4 / 4.9$ & $1.8 / 2.0$ & $0.75$ & $(0.38, 0.29)$ & $(0.21, 0.05)$ & $(0.55, 0.42)$ & $(0.15, 0.07)$ & $(0.39, 0.10)$ & $(0.39, 0.98)$ \\

\midrule
\bottomrule
\end{tabular}
\end{table*}

\subsection{Comparison with only camera-based algorithm}

Table~\ref{table:baseline_comparison} presents a comparative evaluation of the proposed methodology against two camera-only monocular homography baselines across target classes. Both baseline methods (B1 and B2) estimate the 3D centroid and height by projecting upper and lower 2D bounding box points onto the ground plane using state-of-the-art homography, relying on the 2D bounding box bottom-center instead of radar returns. For yaw angle estimation, B1 assumes axis-aligned bounding boxes (no yaw), whereas B2 assigns yaw angles based on road geography look-up tables. 

Across all object classes, the proposed approach significantly outperforms both baselines, particularly in centroid estimation ($x_c, y_c$) and yaw accuracy. For instance, in the car category, the longitudinal position error ($x_c$) is drastically reduced from $(2.67, 2.36)$~m in B1/B2 to $(0.38, 0.29)$~m, while the yaw error is reduced from $(5.88, 4.30)^\circ$ (B2) down to $(3.68, 3.65)^\circ$. Similar substantial gains in positioning accuracy are achieved for pedestrians and bicycles. These improvements are directly attributed to the incorporation of sparse radar point clouds, which provide precise, physically grounded spatial cues that effectively eliminate the severe longitudinal depth ambiguity inherent to purely vision-based homography.

\def\arraystretch{1.1} 
\tabcolsep=12pt 
\begin{table*}[!t]
\centering
\normalsize
\caption{Comparison of bounding box estimation errors (mean, std) with baseline methods (B1: Baseline 1, B2: Baseline 2). Mean and standard deviation are calculated over all $500$ validation frames.}
\label{table:baseline_comparison}
\resizebox{\textwidth}{!}{
\begin{tabular}{c c c c c c c c c}
\toprule \midrule
\textbf{Class} & \textbf{Method} & \textbf{$x_c$ (m)} & \textbf{$y_c$ (m)} & \textbf{$z_c$ (m)} & \textbf{Length (m)} & \textbf{Width (m)} & \textbf{Height (m)} & \textbf{Yaw ($^\circ$)} \\ 
\midrule \hline

\multirow{3}{*}{Person}  
 & B1 & \multirow{2}{*}{$(1.10, 0.89)$} & \multirow{2}{*}{$(0.37, 0.44)$} & \multirow{2}{*}{$(0.21, 0.12)$} & \multirow{2}{*}{$(0.40, 0.09)$} & \multirow{2}{*}{$(0.15, 0.04)$} & \multirow{2}{*}{$(0.27, 0.19)$} & -- \\
 & B2 & & & & & & & -- \\ 
 & \textbf{Ours} & $\mathbf{(0.12, 0.09)}$ & $\mathbf{(0.09, 0.07)}$ & $\mathbf{(0.17, 0.09)}$ & $\mathbf{(0.11, 0.09)}$ & $\mathbf{(0.07, 0.06)}$ & $\mathbf{(0.20, 0.11)}$ & -- \\ 
\hline

\multirow{3}{*}{Bicycle} 
 & B1 & \multirow{2}{*}{$(2.42, 2.43)$} & \multirow{2}{*}{$(0.53, 0.76)$} & \multirow{2}{*}{$(0.14, 0.09)$} & \multirow{2}{*}{$(0.28, 0.11)$} & \multirow{2}{*}{$(0.05, 0.04)$} & \multirow{2}{*}{$(0.16, 0.16)$} & -- \\
 & B2 & & & & & & & $(15.04, 16.27)$ \\ 
 & \textbf{Ours} & $\mathbf{(0.36, 0.22)}$ & $\mathbf{(0.21, 0.16)}$ & $\mathbf{(0.09, 0.05)}$ & $\mathbf{(0.10, 0.09)}$ & $\mathbf{(0.05, 0.04)}$ & $\mathbf{(0.10, 0.08)}$ & $\mathbf{(4.93, 4.63)}$ \\ 
\hline

\multirow{3}{*}{Car}     
 & B1 & \multirow{2}{*}{$(2.67, 2.36)$} & \multirow{2}{*}{$(0.77, 1.27)$} & \multirow{2}{*}{$(0.22, 0.07)$} & \multirow{2}{*}{$(0.21, 0.11)$} & \multirow{2}{*}{$(0.15, 0.07)$} & \multirow{2}{*}{$(0.06, 0.08)$} & -- \\
 & B2 & & & & & & & $(5.88, 4.30)$ \\ 
 & \textbf{Ours} & $\mathbf{(0.38, 0.29)}$ & $\mathbf{(0.55, 0.42)}$ & $\mathbf{(0.22, 0.06)}$ & $\mathbf{(0.21, 0.05)}$ & $\mathbf{(0.15, 0.07)}$ & $\mathbf{(0.08, 0.09)}$ & $\mathbf{(3.68, 3.65)}$ \\

\midrule
\bottomrule
\end{tabular}%
}
\end{table*}
\begin{figure}[!ht]
  \centering
    \begin{subfigure}[b]{0.24\textwidth}
    \centering
    \frame{\includegraphics[width=\textwidth]{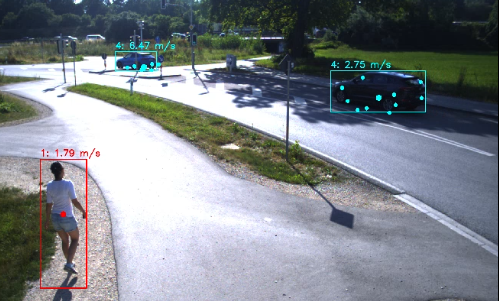}}
    \caption{}
    \label{subfig:a}
  \end{subfigure}
  \begin{subfigure}[b]{0.24\textwidth}
    \centering
    \frame{\includegraphics[width=\textwidth]{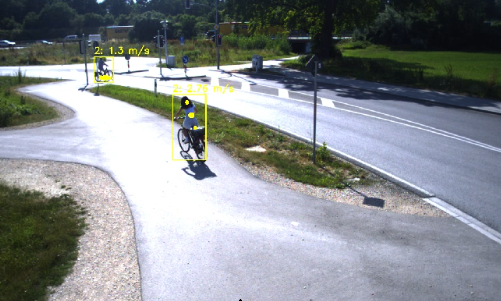}}
    \caption{}
    \label{subfig:b}
  \end{subfigure}
  \vspace{0.2cm}
  \begin{subfigure}[b]{0.24\textwidth}
    \centering
    \frame{\includegraphics[width=\textwidth]{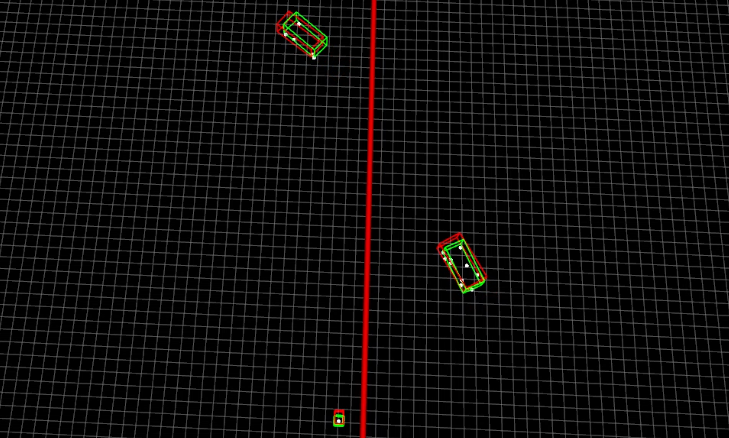}}
    \caption{}
    \label{subfig:c}
  \end{subfigure}
  \begin{subfigure}[b]{0.24\textwidth}
    \centering
    \frame{\includegraphics[width=\textwidth]{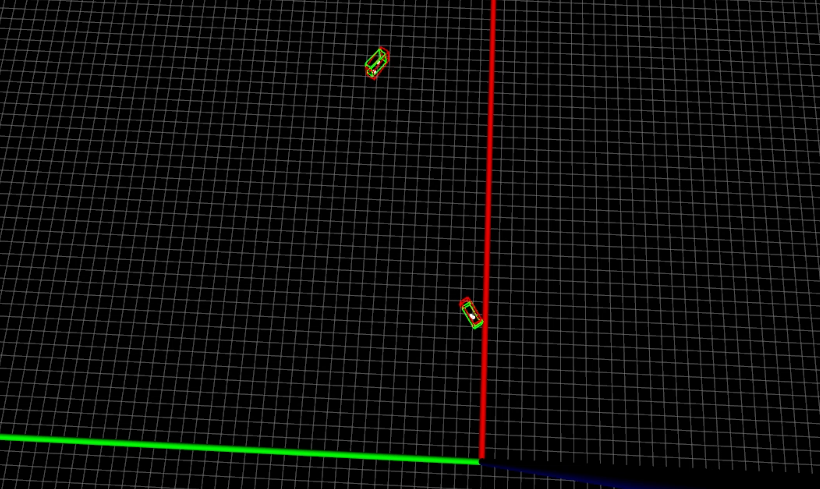}}
    \caption{}
    \label{subfig:d}
  \end{subfigure}
  
  \caption{Output of 3D bounding box estimator. Fig. (a) and (b) shows the image plane with a 2D bounding box, class ID (1:Person, 2:Bicycle, 4:Car), associated radar points, and calculated Doppler velocity, and Fig. (c) and (d) show the corresponding estimated 3D oriented bounding box in the ground plane (green boxes are estimated, red boxes are ground truth, and white spheres are radar points). The red line in (c) and (d) shows the X-axis of the setup.}
  \label{fig:result_images}
\end{figure}

It is worth emphasizing that direct comparisons with end-to-end deep learning camera or fusion frameworks are intentionally omitted, as such models operate under entirely different computational paradigms. Heavy end-to-end architectures require extensive data, high-end hardware, and massive parameter budgets. In contrast, the proposed methodology offers a lightweight, highly efficient, and modular pipeline that executes seamlessly in real time at 25~FPS on an NVIDIA Jetson AGX Orin edge platform, rendering it exceptionally suitable for practical edge deployments where computational resources are tightly constrained. Fig. \ref{fig:result_images} shows some visual examples for reference.

\section{Conclusion} \label{sec_conclusion}

This paper presented a low-cost, resource-efficient two-stage framework for real-time 3D shape and orientation estimation of pedestrians, bicycles, and cars using roadside camera-radar setups. The methodology leverages associated sparse radar point clouds to establish precise 3D target positions, applies bounded dimension templates to stabilize object scale against point sparsity, and employs a lightweight neural network to regress site-specific yaw angles for bicycles and cars accurately. Comprehensive ablation studies established optimal dimension parameters, while comparative evaluations against camera-only homography baselines demonstrated significant performance gains. Operating at $25$~FPS on an NVIDIA Jetson AGX Orin, the proposed system provides a practical, edge-friendly solution for real-time roadside perception.

Despite its efficiency, this work is limited to a single roadside setup, and fixed template bounds cannot capture fine-grained size variations across vehicle sub-classes. Future work will evaluate cross-site transferability across diverse sensor geometries and road geographies, implement dynamic templates, and expand support to heavy vehicles and motorbikes.


\begin{thebibliography}{999}

\bibitem{survey_paper_1}
Z. Wei \emph{et al.}, ``Mmwave radar and vision fusion for object detection in autonomous driving: A review,'' \emph{Sensors}, vol. 22, no. 7, p. 2542, Mar. 2022.

\bibitem{survey_paper_2}
S. Yao \emph{et al.}, ``Radar-camera fusion for object detection and semantic segmentation in autonomous driving: A comprehensive review,'' \emph{IEEE Trans. Intell. Veh.}, vol. 9, no. 1, pp. 2094--2128, Jan. 2024.

\bibitem{shiva_vehits_2022} 
S. Agrawal \emph{et al.}, ``Concept of smart infrastructure for connected vehicle assist and traffic flow optimization,'' in \emph{Proc. 8th Int. Conf. Vehicle Technol. Intell. Transp. Syst. (VEHITS)}, Online Event, Apr. 26--28, 2022, pp. 360--367.

\bibitem{in2lab_shiva} 
S. Agrawal and G. Elger, ``Concept of infrastructure based environment perception for IN2Lab test field for automated driving,'' in \emph{Proc. IEEE Int. Smart Cities Conf. (ISC2)}, Manchester, UK, Sep. 7--10, 2021, pp. 1--4.

\bibitem{infra_low_cr_fusion_paper_1}
K. Li, H. Cui, Z. Dai, and H. Song, ``RVIFNet: Radar-visual information fusion network for all-weather vehicle perception in roadside monitoring scenarios,'' \emph{IEEE Sensors J.}, vol. 24, no. 23, pp. 40105--40122, Dec. 2024.

\bibitem{infra_low_cr_fusion_paper_2}
Y. Li \emph{et al.}, ``FARFusion V2: A geometry-based radar-camera fusion method on the ground for roadside far-range 3D object detection,'' in \emph{Proc. 32nd ACM Int. Conf. Multimedia (MM)}, 2024, pp. 8421--8430.

\bibitem{rc_high_level_vehicle_1}
Z. Zhong, S. Liu, M. Mathew, and A. Dubey, ``Camera radar fusion for increased reliability in ADAS applications,'' \emph{Electron. Imaging}, vol. 30, pp. 1--4, Jan. 2018.

\bibitem{lidar_3d_object_detection_survey_1}
G. Zamanakos, L. Tsochatzidis, A. Amanatiadis, and I. Pratikakis, ``A comprehensive survey of LIDAR-based 3D object detection methods with deep learning for autonomous driving,'' \emph{Comput. Graph.}, vol. 99, pp. 153--181, Oct. 2021.

\bibitem{lidar_3d_object_detection_survey_2}
S. Y. Alaba and J. E. Ball, ``A survey on deep-learning-based LiDAR 3D object detection for autonomous driving,'' \emph{Sensors}, vol. 22, no. 24, Art. no. 9577, Dec. 2022.

\bibitem{mono_camera_3d_object_detection_infra_1}
L. Yang \emph{et al.}, ``MonoGAE: Roadside monocular 3D object detection with ground-aware embeddings,'' \emph{IEEE Trans. Intell. Transp. Syst.}, vol. 25, no. 11, pp. 17587--17601, Nov. 2024.

\bibitem{mono_camera_3d_object_detection_infra_3}
G. Miao, R. Quan, C. Pan, Z. Hu, and J. Qin, ``BEVTemp: Enhancing vision-based roadside 3D object detection with temporal information,'' in \emph{Proc. Pacific Rim Int. Conf. Artif. Intell. (PRICAI)}, 2025, pp. 233--245.

\bibitem{mono_camera_3d_object_detection_infra_4}
Z. Ye, H. Zhang, J. Gu, and X. Li, ``Yolov7-3d: A monocular 3d traffic object detection method from a roadside perspective,'' \emph{Appl. Sci.}, vol. 13, no. 20, Art. no. 11402, Oct. 2023.

\bibitem{cr_low_only_cars_3d_output_paper_1}
Y. Liu, S. Chang, Z. Wei, K. Zhang, and Z. Feng, ``Fusing mmWave radar with camera for 3-D detection in autonomous driving,'' \emph{IEEE Internet Things J.}, vol. 9, no. 20, pp. 20408--20421, Oct. 2022.

\bibitem{cr_low_vehicle_3d_output_paper_1}
R. Nabati and H. Qi, ``Centerfusion: Center-based radar and camera fusion for 3d object detection,'' in \emph{Proc. IEEE/CVF Winter Conf. Appl. Comput. Vis. (WACV)}, 2021, pp. 1527--1536.

\bibitem{vehicle_3d_fusion_he2023concs}
W. He, Z. Deng, Y. Ye, and P. Pan, ``ConCs-Fusion: A context clustering-based radar and camera fusion for three-dimensional object detection,'' \emph{Remote Sens.}, vol. 15, no. 21, Art. no. 5130, Nov. 2023.

\bibitem{cr_low_vehicle_3d_output_paper_4}
P. Kalgaonkar and M. El-Sharkawy, ``NeXtFusion: Attention-based camera-radar fusion network for improved three-dimensional object detection and tracking,'' \emph{Future Internet}, vol. 16, no. 4, Art. no. 114, Apr. 2024.

\bibitem{C01}
L. Blumenthal, ``Real-Time Monocular 3D Object Detection to Support Autonomous Driving,'' Bachelor's thesis, Tech. Univ. Munich, Munich, Germany, 2022.

\bibitem{C02}
F. Fraundorfer, P. M. Roth, and F. Schenk, ``Perspective transformation for accurate detection of 3D bounding boxes of vehicles in traffic surveillance,'' in \emph{Proc. 24th Comput. Vis. Winter Workshop (CVWW)}, 2019, pp. 1--9.

\bibitem{C03}
M. H. Zwemer, D. Scholte, R. G. J. Wijnhoven, and P. H. N. de With, ``3D detection of vehicles from 2D images in traffic surveillance,'' in \emph{Proc. 17th Int. Joint Conf. Comput. Vis., Imag. Comput. Graph. Theory Appl. (VISIGRAPP - VISAPP)}, 2022, pp. 97--106.

\bibitem{C04}
A. Mousavian, D. Anguelov, J. Flynn, and J. Kosecka, ``3D bounding box estimation using deep learning and geometry,'' in \emph{Proc. IEEE Conf. Comput. Vis. Pattern Recognit. (CVPR)}, 2017, pp. 7074--7082.

\bibitem{C05}
M. Rezaei, M. Azarmi, and F. M. P. Mir, ``3D-Net: Monocular 3D object recognition for traffic monitoring,'' \emph{Expert Syst. Appl.}, vol. 227, Art. no. 120253, Oct. 2023, doi: 10.1016/j.eswa.2023.120253.

\bibitem{R01}
Y. Feng, J. Zhao, C. Wang, L. Xie, and S. Lu, ``3D bounding box estimation based on COTS mmWave radar via moving scanning,'' \emph{Proc. ACM Interact. Mob. Wearable Ubiquitous Technol.}, vol. 8, no. 4, pp. 1--27, Dec. 2024.

\bibitem{R02}
Y. Lee, J. Kim, S. Kim, H. Lee, and S. Lee, ``Estimation of moving direction and size of vehicle in high-resolution automotive radar system,'' \emph{IEEE Trans. Intell. Transp. Syst.}, vol. 25, no. 7, pp. 7174--7186, Jul. 2024.

\bibitem{R03}
K. Bansal, K. Rungta, S. Zhu, and D. Bharadia, ``Pointillism: Accurate 3D bounding box estimation with multi-radars,'' in \emph{Proc. 18th ACM Conf. Embedded Netw. Sensor Syst. (SenSys)}, 2020, pp. 340--353.

\bibitem{R04}
F. Roos, D. Kellner, J. Dickmann, and C. Waldschmidt, ``Reliable orientation estimation of vehicles in high-resolution radar images,'' \emph{IEEE Trans. Microw. Theory Techn.}, vol. 64, no. 9, pp. 2986--2993, Sep. 2016.

\bibitem{R05}
J. Schlichenmaier, N. Selvaraj, M. Stolz, and C. Waldschmidt, ``Template matching for radar-based orientation and position estimation in automotive scenarios,'' in \emph{Proc. IEEE MTT-S Int. Conf. Microw. Intell. Mobil. (ICMIM)}, 2017, pp. 95--98.

\bibitem{L01}
J. Chen, M. Song, and J. Wang, ``Efficient L-shape fitting based on critical edge for vehicle orientation estimation using LiDAR,'' in \emph{Proc. IAVSD Int. Symp. Dyn. Veh. Roads Tracks}, 2024, pp. 283--292.

\bibitem{L02}
T. Nystr{\"o}m Johansson and O. Wellenstam, ``LiDAR clustering and shape extraction for automotive applications,'' Master's thesis, Dept. Signals Syst., Chalmers Univ. Technol., Gothenburg, Sweden, 2017.

\bibitem{L03}
C. Zhao, C. Fu, J. M. Dolan, and J. Wang, ``L-shape fitting-based vehicle pose estimation and tracking using 3D-LiDAR,'' \emph{IEEE Trans. Intell. Veh.}, vol. 6, no. 4, pp. 787--798, Dec. 2021.

\bibitem{sensor_fusion_mdpi_shiva} 
S. Agrawal, S. Bhanderi, and G. Elger, ``Infra-3DRC-FusionNet: Deep fusion of roadside mounted RGB mono camera and three-dimensional automotive radar for traffic user detection,'' \emph{Sensors}, vol. 25, no. 11, p. 3422, May 2025.

\bibitem{shiva_ccis_springer_paper} 
S. Agrawal, R. Song, K. Doycheva, A. Knoll, and G. Elger, ``Intelligent roadside infrastructure for connected mobility,'' in \emph{Smart Cities, Green Technologies, and Intelligent Transport Systems} (Communications in Computer and Information Science, vol. 1843), C. Klein, M. Jarke, J. Ploeg, M. Helfert, K. Berns, and O. Gusikhin, Eds. Cham, Switzerland: Springer, Jul. 2023, pp. 134--157.

\bibitem{camera_intrinsic_checkerboard}
Z. Zhang, ``A flexible new technique for camera calibration,'' \emph{IEEE Trans. Pattern Anal. Mach. Intell.}, vol. 22, no. 11, pp. 1330--1334, Nov. 2000.

\bibitem{shiva_auto_calibration_paper} 
S. Agrawal, S. Bhanderi, K. Doycheva, and G. Elger, ``Static multi-target-based auto-calibration of RGB cameras, 3D Radar, and 3D Lidar sensors,'' \emph{IEEE Sensors J.}, vol. 23, no. 18, pp. 21493--21505, Sept. 2023.

\bibitem{cr_high_infra_3}
Y. Fu \emph{et al.}, ``A camera-radar fusion method based on edge computing,'' in \emph{Proc. IEEE Int. Conf. Edge Comput. (EDGE)}, 2020, pp. 9--14.

\bibitem{cr_high_infra_4}
F. A. Jibrin, Z. Deng, and Y. Zhang, ``An object detection and classification method using radar and camera data fusion,'' in \emph{Proc. IEEE Int. Conf. Signal, Inf. Data Process. (ICSIDP)}, 2019, pp. 1--6.

\bibitem{yolo11_ultralytics}
G.~Jocher and J.~Qiu, ``Ultralytics YOLO11,'' ver. 11.0.0, 2024. [Online]. Available: \url{https://github.com/ultralytics/ultralytics}

\bibitem{savan_radar_paper} 
S. Bhanderi, S. Agrawal, and G. Elger, ``Deep segmentation of 3+1D radar point cloud for real-time roadside traffic user detection,'' \emph{Sci. Rep.}, vol. 15, no. 1, p. 38489, Nov. 2025.

\end{thebibliography}
\end{document}